\documentclass[twocolumn,9pt]{article}

\usepackage{wrapfig}
\usepackage{booktabs}
\usepackage{geometry}
\usepackage{multirow}
\usepackage[utf8]{inputenc}
\newcommand{\footremember}[2]{%
    \footnote{#2}
    \newcounter{#1}
    \setcounter{#1}{\value{footnote}}%
}
\newcommand{\footrecall}[1]{%
    \footnotemark[\value{#1}]%
} 

\title{Google Landmark Recognition 2020 Competition Third Place Solution}
\author{%
  Qishen Ha \footremember{equal}{Equal contribution} \\ \small LINE Corp%
  \and 
  Bo Liu\footrecall{equal} \footnote{Corresponding author. Email: boli@nvidia.com} \\ \small NVIDIA%
  \and 
  Fuxu Liu\footrecall{equal} \\ \small ReadSense Ltd%
  \and
  Peiyuan Liao\footrecall{equal} \\ \small Carnegie Mellon University%
  }

\date{\today}

\usepackage{natbib}
\usepackage{graphicx}

\begin{document}
\maketitle

\begin{abstract}
We present our third place solution to the Google Landmark Recognition 2020 competition. It is an ensemble of global features only Sub-center ArcFace models. We introduce dynamic margins for ArcFace loss, a family of tune-able margin functions of class size, designed to deal with the extreme imbalance in GLDv2 dataset. Progressive finetuning and careful postprocessing are also key to the solution. Our two submissions scored 0.6344 and 0.6289 on private leaderboard, both ranking third place out of 736 teams. 
\end{abstract}

\section{Introduction}

Google Landmark Retrieval and Recognition competitions have been running for the third consecutive year. In previous years, winning solutions used a combination of global and local features, such as last year's first place winners in recognition \cite{2019recognition1st} and retrieval \cite{2019retrieval1st} tracks. 

Interestingly, the advances in state-of-the-art image classification models over the past year, especially EfficientNet \cite{efficientnet}, together with their open source ImageNet pretrained weights, have pushed the performance of global feature only models on this task so high that the additional benefit of local feature models could be marginal or negligible. This is evidenced by the fact that none of top three solutions in Retrieval 2020 \cite{Retrieval1st,Retrieval2nd,Retrieval3rd} nor the first place solution in Recognition 2020 \cite{2020recog1st} used local features.

Our solution, too, consists of only global feature models. It is an ensemble of 7 metric learning models. The architecture of choice is Sub-center ArcFace \cite{subcenter}. To alleviate the extreme class imbalance in the GLDv2 dataset \cite{GLDv2}, we propose dynamic margins for ArcFace loss, a family of continuous functions of class size, which gives major performance boost over constant margins. We designed a progressive finetuning schedule to deal with the huge dataset in limited time. Lastly, we applied postprocessing strategies taking advantage of both feature similarity search and ArcFace head's output. We attempted to use local feature models like DELG \cite{Delg}, but they did not boost the ensemble's performance.

In section 2, we explain the dynamic margins in more details. Training schedule and setup are discussed in section 3. We then describe our test set predicting strategy in section 4. Finally, the ensemble and scores are presented in section 5.

\section{Model architecture: Sub-center ArcFace with dynamic margins}

ArcFace \cite{arcface} has been the state-of-the-art metric learning methods. In this competition, we used Sub-center ArcFace \cite{subcenter}, a recent improvement over ArcFace, designed to better handle noisy data. Sub-center ArcFace may be particularly suitable for GLDv2 dataset besides the fact that it is noisy. The idea is that each class may have more than one class centers. For example, a certain landmark's photos may have a few clusters (e.g. taken from different angles). Sub-center ArcFace's model weights include multiple class centers' representations, which can increase classification accuracy and improve global feature's quality. We chose number of sub-centers $K=3$.

Another feature of the GLDv2 dataset is that it's extremely imbalanced with very longs tails. Inspired by AdaptiveFace \cite{adaptiveface}, we aim to assign different margin levels for different class sizes. For models to converge better in the presence of heavy imbalance, smaller classes need to have bigger margins as they are harder to learn.

\begin{figure}
\includegraphics[width=0.9\linewidth]{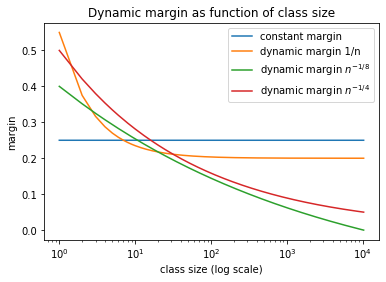}
\caption{\textbf{Constant margin and three dynamic margins as function of class size} }
\label{fig:model}
\end{figure}

Instead of manually setting different margin levels based on class size, we introduce dynamic margins, a family of continuous functions mapping class size to margin level: $$f(n) = a\cdot n^{-\lambda}+b$$
where $a,b,\lambda$ are parameters. $\lambda>0$ determines the shape of the margin function. The closer $\lambda$ is to 0, the similar it is to a linear function. $a$ and $b$ determines the upper of lower bounds of each margin function. See Figure \ref{fig:model}.

We experimented with $\lambda=1, \frac14, \frac18$ using a baseline EfficientNet-B0 model with image size 256. We tuned the optimal lower and upper bounds of each function separately. The model score of each dynamic function are listed in Table \ref{tab:margin}, compared with constant margin at its optimal level 0.25. 

The best dynamic function is $n^{-1/4}$ with lower and upper bounds 0.05 and 0.5 respectively. It has over 0.025 gain in validation GAP score over constant margin baseline. We used this best dynamic function to train all our models in the final ensemble.

\begin{table*}[h]
\begin{center}
\begin{tabular}{cccc}
\toprule
margin function   & lower and upper bounds   & validation accuracy & validation GAP \\
\midrule
constant margin& 0.25 & 0.4762           & 0.84176       \\
1/n & 0.2 -- 0.55      & 0.4805           & 0.85475       \\
$n^{-1/8}$& 0 -- 0.4   & 0.4814           & 0.86471       \\
$n^{-1/4}$& 0.05 -- 0.5 & \textbf{0.4822}           & \textbf{0.86710}    \\ 
\bottomrule                     
\end{tabular}
\caption{Validation scores of three dynamic margin functions of baseline EfficientNet B0 model with 256 image size, compared with the constant margin counterpart. The constant margin level 0.25 and the upper and lower bounds of each dynamic margin functions are tuned optimal values. }\label{tab:margin}
\end{center}
\end{table*}

\section{Training schedule and setup}

In the competition dataset, cGLDv2 (cleaned GLDv2), there are 1.6 million images and 81k classes. All landmark test images belong to these classes. In the original GLDv2 dataset\cite{GLDv2}, there are 4.1 million images and 203k classes, among which 3.2 million images belong to the 81k cGLDv2 classes. 

We noticed that (1) training with the 3.2 million data gives better results than only the 1.6 million competition cGLDv2 data, (2) pretraining on all 4.1 million data then finetuning on 3.2 million data gives even better results.

Therefore, we designed the following progressive finetuning strategy similar to the winning solution in Google Landmark Retrieval 2020 \cite{Retrieval1st}:
\begin{itemize}
    \item Stage 1 (pretrain): 10 epochs with small image size (256) on the full 4.1 million images
    \item Stage 2 (finetune): 15-20 epochs with medium image size (512 to 768 depending on model) on 3.2 million images
    \item Stage 3 (finetune): 1-10 epoch with large image size (672 to 1024) on 3.2 million images
\end{itemize}

We used the following augmentations during training: horizontal flip, image compression, shift, scale, rotate and cutout\cite{cutout}. We find heavier augmentations hurt model performance. All images are resized into square shape without cropping.

Training data is split into stratified 5-fold. Each model is trained on 4 folds and validated on only 1/15 of the other fold in order to save time. We used different folds to train different single models to increase model diversity.

GAP metric is used for validation scores, same as the leaderboard. But validation scores are much higher than leaderboard scores, due to the absence of non-landmark images in our validation set. All our single model's validation GAP scores are over 0.967.

\section{Test set predicting strategy}

\subsection{Baseline approach}
The ArcFace model can output a 512-dimensional global feature for any input image. For each image in the query set (i.e. test set), calculate its global feature cosine similarity between all the gallery set (i.e. private train set) images. Use the top 1 nearest neighbor's class and the corresponding cosine similarity as the prediction and confidence score for this query image.

Our best single model scored \textbf{0.5859/0.6040} (private/public) using this baseline approach.

\subsection{Postprocessing step 1}
Instead of using top 1 nearest neighbor only, we can improve it by combining the same-class images in top 5 nearest neighbors and their cosine similarities. We found the best combining function to be 8th power, i.e. applying the 8th power function to each individual scores before summing. A high power ensures that high values in cosine similarity dominate lower values, making it harder for top 1 predictions to be overturned.

\begin{figure*}[]
\centering
\includegraphics[width=1\textwidth]{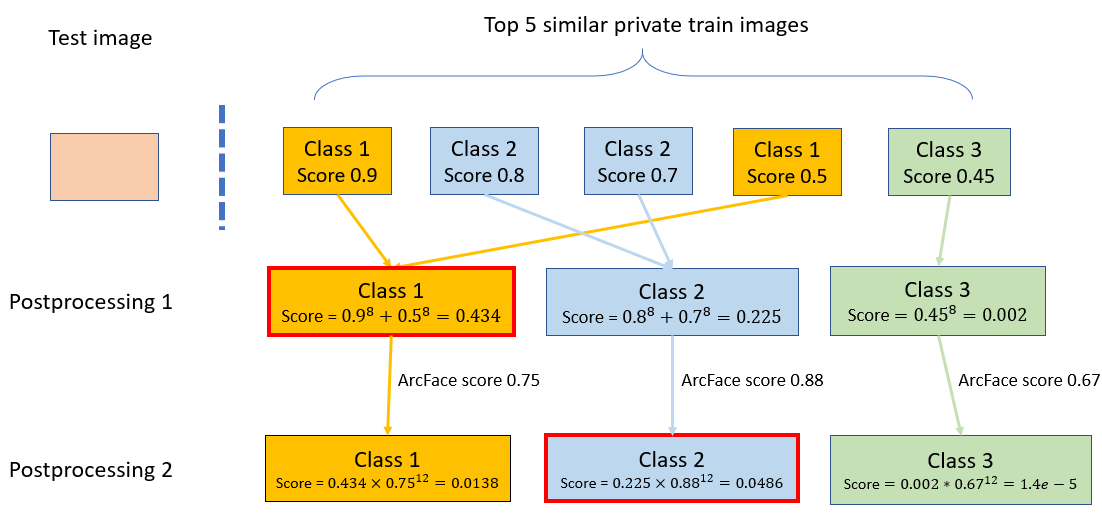}
\caption{Postprocessing illustration.}
\label{fig:PP}
\end{figure*}

It is illustrated in Fig \ref{fig:PP}. In this example, 1st and 4th nearest neighbors are the same class while the 2nd and 3rd nearest neighbors are the same class. After combining, the predicted class is still class 1. But the score changed, which has an effect on the relative position of this test image in GAP metric calculation.

Best single model's score improved to \textbf{0.5932/0.6104} (private/public) after postprocessing step 1.

\subsection{Postprocessing step 2}
We can further improve it by incorporating ArcFace head's predicted scores. They represent the test image's cosine similarity to each class center, which contain information the model learned from the whole public train set. The best function to incorporate ArcFace head's score is found to be 12th power.

Postprocessing step 2 is also illustrated in Fig \ref{fig:PP}. In this example, class 2's ArcFace head score is higher than class 1, boosting class 2 to have the highest overall score.

Best single model's score increased to \textbf{0.6014/0.6244} (private/public) with both postprocessing steps.

\section{Ensemble}

We trained 7 models for the ensembles: EfficientNets-B7, B6, B5, B4, B3\cite{efficientnet}, ResNeSt-101\cite{resnest}, ReXNet2.0\cite{rexnet}.

In ensemble 1, each model is pretrained for 10 epochs on full GLDv2 data at small image size (256), then finetuned on the 81313 subset of GLDv2 data at medium image size (512 to 768) for 13-21 epochs, then finetuned again on the same data at large image size (672 to 1024) for 1 epoch. See Table \ref{tab:score1} for detailed model configuration and scores.

In order to have diversity between the two submissions, we finetuned the models for more epochs in ensemble 2. In some models, the finetune image size also changed. Moreover, a second Efficient-B5 and B6 models (the two best single models) are added to ensemble 2, making it a 9-model ensemble. Detailed model configuration and scores are listed in Table \ref{tab:score2}. 

For global feature neighbor search, we concatenated each model's 512-dimensional feature before computing the cosine similarities between gallery and query images; for ArcFace head, we take simple average of each model's output.

Ensemble 1 has higher public LB score (0.6604), but ensemble 2 has higher private LB score (0.6344). Both submissions are good enough for third place on private leaderboard.

\begin{table*}[ht]
\begin{center}
\small
\setlength\tabcolsep{1.5pt} 
\begin{tabular}{p{2.7cm}p{2.3cm}p{1.5cm}cc}
\toprule
Model        &Image Size&	Epochs	&Private LB	&Public LB\\
\hline
EfficientNet-B7	&256, 512, 672	&10, 13, 1	&0.5678 & 0.5838\\
EfficientNet-B6	&256, 512, 768	&10, 17, 1	&0.5807 & 0.6030\\	
EfficientNet-B5	&256, 576, 768	&10, 16, 1	&0.5829	& 0.6037\\
EfficientNet-B4	&256, 704, 768	&10, 16, 1	&0.5768 & 0.5905\\
EfficientNet-B3	&256, 544, 1024	&10, 18, 1	&0.5687 & 0.5857\\	
ResNeSt101      &256, 576, 768	&10, 16, 1	&0.5711 & 0.5953\\	
ReXNet2.0	    &256, 768, 1024	&10, 21, 1	&0.5757 & 0.5998\\
\hline
Ensemble			&&&0.6109&	0.6368\\
Ensemble with PP		&&&\textbf{0.6289}&	0.6604\\
\bottomrule
\end{tabular}
\caption{\textbf{Ensemble 1's model configuration and scores.} The three numbers in Image Size and Epochs columns are the image size and number of epochs in each of the three training stages. Single model's scores are without postprocessing (PP).}\label{tab:score1}
\end{center}
\end{table*}

\begin{table*}[ht]
\begin{center}
\small
\setlength\tabcolsep{1.5pt} 
\begin{tabular}{p{2.7cm}p{2.3cm}p{1.5cm}cc}
\toprule
Model &	Image Size&	Epochs	&Private LB	&Public LB\\
\hline
EfficientNet-B7	&256, 512, 672	&10, 13, 5		\\
EfficientNet-B6	&256, 512	&10, 36	&0.5842	&0.6028\\
EfficientNet-B6	&256, 512, 768	&10, 28, 5	\\	
EfficientNet-B5	&256, 576, 768	&10, 16, 1	&0.5829	&0.6037\\
EfficientNet-B5	&256, 576, 768	&10, 33, 4		\\
EfficientNet-B4	&256, 704, 768	&10, 16, 5	&0.5748	&0.5944\\
EfficientNet-B3	&256, 544, 768	&10, 29, 10	\\	
ResNeSt101&	256, 576, 768	&10, 16, 6	\\	
ReXNet2.0	&256, 768	&10, 38		\\
\hline
Ensemble			&&& 0.6134 & 0.6318\\
Ensemble with PP		&&&\textbf{0.6344}&	0.6581\\
\bottomrule
\end{tabular}
\caption{\textbf{Ensemble 2's model configuration and scores.} The numbers in Image Size and Epochs columns are the image size and number of epochs in each of the two or three training stages. Empty single model score means we did not submit those epochs individually. There are two more single models in ensemble 2 than in ensemble 1.}\label{tab:score2}
\end{center}
\end{table*}

\bibliographystyle{plain}
\bibliography{references}
\end{document}